\documentclass{article}

\usepackage{times}
\usepackage{graphicx} 
\usepackage{subfigure} 
\usepackage{amsmath}
\usepackage{amssymb}
\usepackage{wrapfig}

\usepackage{natbib}

\usepackage{algorithm}
\usepackage{algorithmic}

\usepackage[accepted]{icml2014}

\icmltitlerunning{Learning to encode motion using spatio-temporal synchrony}

\begin{document} 

\twocolumn[
\icmltitle{Learning to encode motion using spatio-temporal synchrony}

\icmlauthor{Kishore Konda}{konda@informatik.uni-frankfurt.de}
\icmladdress{Goethe University Frankfurt,
            Frankfurt}
\icmlauthor{Roland Memisevic}{roland.memisevic@umontreal.ca}
\icmladdress{University of Montreal,
            Montreal}
\icmlauthor{Vincent Michalski}{vmichals@rz.uni-frankfurt.de}
\icmladdress{Goethe University Frankfurt,
            Frankfurt}    

\icmlkeywords{motion estimation, machine learning, activity recognition}

\vskip 0.3in
]

\begin{abstract} 
We consider the task of learning to extract motion from videos. 
To this end, we show that the detection of spatial transformations 
can be viewed as the detection of synchrony between the image sequence 
and a sequence of features undergoing the motion we wish to detect. 
We show that learning about synchrony is possible using very fast, local 
learning rules, by introducing multiplicative ``gating'' interactions between
hidden units across frames. 
This makes it possible to achieve competitive performance in 
a wide variety of motion estimation tasks, using a small fraction of the 
time required to learn features, and to outperform 
hand-crafted spatio-temporal features by a large margin. 
We also show how learning about synchrony can be viewed as performing greedy
parameter estimation in the well-known motion energy model. 
\end{abstract} 

\section{Introduction}

The classic motion energy model turns the frames of a video into a representation 
of motion by summing over squares of Gabor filter responses \cite{Adelson85,Watson:85}. 
One of the motivations for this computation is the fact that sums over squared 
filter responses allow us to detect ``oriented'' energies in spatio-temporal frequency bands. 
This, in turn, makes it possible to encode motion independently of phase information, 
and thus to represent motion to some degree independent of what is moving. 
Related models have been proposed for binocular disparity estimation 
\citep[e.g.][]{FleetBinocular},
which also involves the 
estimation of the displacement of local features across multiple views. 

For many years, hand-crafted, Gabor-like filters have been used 
\citep[see, e.g.,][]{YUPENN},
but 
in recent years, unsupervised deep learning techniques 
have become popular which learn the features 
from videos 
\citep[e.g.][]{Taylor:2010,ISA,3DCNN,Memisevic07},
The interest in learning-based models of motion 
is fueled in part by the observation that for activity recognition, 
hand-crafted features tend to not perform uniformly well across 
tasks \cite{Wang09evaluationof}, which suggests learning the features 
instead of designing them by hand.

Unlike images, videos have been somewhat resistant to feature learning, in that 
many standard models do not work well. 
On images, for example, models like the autoencoder or even K-means clustering, 
are known to yield highly structured, Gabor-like, filters, which perform well in 
recognition tasks \cite{coatessinglelayer}. 
The same does not seem to be true for videos, where neither autoencoders nor K-means were shown to 
work well (see, for example, Section~\ref{section:experiments}). 
There are two notable exceptions: Feature learning models like ICA, where inference 
involves a search over filters that are sparse and at the same time minimize squared 
reconstruction error, were shown to learn at least visually good filters
(see, for example, \cite{OlshausenTimevarying} and references in \cite{Hyvarinen}, Chapter 16). 
The other exception are energy models, which 
compute sums over squared filter responses for inference, and  
which were shown to work well in activity recognition tasks \citep[e.g.][]{ISA}. 

In this work, we propose a possible explanation for why some models work well 
on videos, and other models do not. 
We show that a linear encoding permits the detection of transformations across time, 
because it supports the detection of temporal ``synchrony'' between video and features. 
This makes it possible to interpret motion energy models as a way to combine 
two independent contributions to motion encoding, namely the detection of 
synchrony, and the encoding of invariance. 
We show how disentangling these two contributions provides a different perspective 
onto the energy model and suggests new approaches to learning. 
In particular, we show that learning a linear encoding can be viewed as learning in the 
presence of multiplicative ``gating'' interactions \citep[e.g.][]{Mel94}.
This allows us to learn competitive motion features on conventional CPU-based hardware and in 
a small fraction of the time required by previous methods.


\section{Motion from spatio-temporal synchrony}
\label{section:synchrony}
Consider the task of computing a representation of motion, given two frames $\vec{x}_1$ and $\vec{x}_2$ of a video. 
The classic energy model \cite{Adelson85} solves this task by detecting subspace energy. 
This amounts to computing the sum of squared quadrature Fourier or Gabor coefficients 
across multiple frequencies and orientations \citep[e.g.][]{Hyvarinen}. 
The motivation behind the energy model is that Fourier amplitudes are independent of 
stimulus phase, so they yield a representation of motion that is to some degree independent 
of image content. As we shall show below, this view confounds two independent contributions 
of the energy model, which may be disentangled in practice.  

An alternative to computing the sum over \emph{squares}, which has originally been 
proposed for stereopsis, is the cross-correlation model \cite{arndt1995human,FleetBinocular},  
which computes the sum over \emph{products} of filter-responses across the 
two frames. 
It can be shown that the sum over products of filter responses in quadrature 
encodes angles in the invariant subspaces associated with the transformation. 
The representation of angles thereby also yields a phase-invariant representation 
of motion \citep[e.g.][]{FleetBinocular,OlshausesCadieu,multiview}. 
Like the energy model, it also confounds invariance and representing transformations
as we shall show. 

It can be shown that cross-correlation models and energy models are closely related,  
and that there is a canonical operation that turns one into the other 
\citep[e.g.][]{FleetBinocular,multiview}.
We shall revisit the close relationship between these models
in Section~\ref{subsection:evensymmetric}.

\subsection{Motion estimation by synchrony detection}
\label{subsection:synchrony_detection}
We shall now discuss how synchrony detection allows us 
to compute motion, and how content-invariance can be achieved by pooling afterwards, if desired. 
To this end, consider two filters $\vec{w}_1$ and $\vec{w}_2$ which shall encode the transformation between two 
images $\vec{x}_1$ and $\vec{x}_2$. 
We restrict our attention to transformations that can be represented as an orthogonal 
transformation in ``pixel space'', in other words, as an orthogonal image warp. 
As these include all permutations, they include, in particular,
most common spatial transformations such as local translations and their combinations 
\citep[see, e.g.][for a recent discussion]{multiview}.
The assumption of orthogonality of transformations is made implicitly also by 
the motion energy model.

To detect an orthogonal transformation, $P$, between the two images, we propose to use 
filters for which 
\begin{equation}
\label{eq:synchronyfilters}
\vec{w}_2 = P \vec{w}_1
\end{equation}
holds, and then to check whether the condition 
\begin{equation} 
\label{eq:synchrony}
\vec{w}_2^\mathrm{T} \vec{x}_2
=
\vec{w}_1^\mathrm{T} \vec{x}_1 
\end{equation}
is true. 
We shall call this the "synchrony condition".
It amounts to choosing a filter pair, such that it is an example of the transformation we want 
to detect (Eq.~\ref{eq:synchronyfilters}), and to determine whether the two 
filters yield equal responses when applied in sequence to the two 
frames (Eq.~\ref{eq:synchrony}). 
We shall later relax the exact equality to an approximate equality.  

To see why the synchrony condition counts as evidence for the presence of the 
transformation, note first that $\vec{x}_2 = P \vec{x}_1$ implies $\vec{w}_2^\mathrm{T} \vec{x}_2 = \vec{w}_2^\mathrm{T} P \vec{x}_1$.

From this, we get: 
\begin{eqnarray}
& \vec{x}_2 = P \vec{x}_1 \; (\text{presence of $P$}) \nonumber \\
\Rightarrow & \; \vec{w}_2^\mathrm{T} \vec{x}_2 \;\Big( =  \vec{w}_2^\mathrm{T} P \vec{x}_1 =  (P^\mathrm{T} \vec{w}_2)^\mathrm{T} \vec{x}_1 = \Big) \; \vec{w}_1^\mathrm{T} \vec{x}_1 
\end{eqnarray}
The last equation follows from $P^\mathrm{T}=P^\mathrm{-1}$ (orthogonality of $P$).
This shows that the presence of the transformation $P$ implies synchrony (Eq.\ref{eq:synchrony}) 
for any two filters which themselves are related through $P$, 
that is $\vec{w}_2=P\vec{w}_1$. 
In order to detect the presence of $P$, we may thus look for the synchrony condition, using 
a set of filters transformed through $P$. 
This is an inductive (statistical) reasoning step, in that we can accumulate evidence 
for a transformation by looking for synchrony across multiple filters. 
The absence of the transformation implies that all filter pairs violate the synchrony condition. 

It is interesting to note that for Gabor filters, phase shifts and position shifts are 
locally the same 
\citep[e.g.][]{FleetBinocular}.
For global Fourier features, phase shifts and position shifts are exactly identical. 
Thus, synchrony (Eq.~\ref{eq:synchronyfilters}) between the inputs and a sequence 
of phase-shifted Fourier (or Gabor) features, for example, allows us to detect 
transformations which are \emph{local translations}. 
We shall discuss learning of filters from video data in Section~\ref{section:model}.

The synchrony condition can be extended to a sequence of more than two 
frames as follows: Let $\vec{x}_i, \vec{w}_i \; (i=1,\dots,T)$ denote the input 
frames and corresponding filters. 
To detect a set of transformations $P_{i}$, each of which relates two adjacent frames 
$(\vec{x}_i, \vec{x}_{i+1})$, 
set $\vec{w}_{i+1} = P_i \vec{w}_i$ for all $i$. 
The condition for the presence of the sequence of transformations now turns into 
\begin{equation}
\label{synchrony_seq}
\vec{w}_i^\mathrm{T} \vec{x}_i = \vec{w}_j^\mathrm{T} \vec{x}_j \quad \forall i,j=1,\dots,T \text{ and } i\neq j
\end{equation}

\subsection{The insufficiency of weighted summation}
To check for the synchrony condition in practice, it is necessary 
to detect the equality of transformed filter responses across time (Eq.~\ref{eq:synchrony}). 
Most current deep learning models are based on layers of weighted summation followed by 
a nonlinearity.
The detection of synchrony, unfortunately, cannot be performed in a layer of 
weighted summation plus nonlinearity 
as we shall discuss now. 

The fact that the sum of filter responses, $\vec{w}_1^\mathrm{T}\vec{x}_1 + \vec{w}_2^\mathrm{T}\vec{x}_2$,  
will attain its maximum for inputs that both match their filters seems to suggest 
that thresholding it would allow us to detect synchrony. 
This is not the case, however, because thresholding works well only for 
inputs which are very similar to the feature vectors themselves: 
Most inputs, in practice, will be normalized superpositions of \emph{multiple} feature vectors. 
Thus, to detect synchrony with a thresholded sum, 
we would need to use a threshold small enough to represent 
features, $\vec{w}_1$, $\vec{w}_2$, 
that explain only a fraction of the variability in $\vec{x}_1, \vec{x}_2$. 
If we assume, for example, that the two features $\vec{w}_1$, $\vec{w}_2$ 
account for $50\%$ of the variance in the inputs (an overly optimistic assumption), 
then we would have to reduce the threshold to be one half of the maximum attainable 
response to be able to detect synchrony. 
However, at this level, there is no way to distinguish between two stimuli which 
do satisfy the synchrony condition (\emph{the motion in question is present}), 
and two stimuli where one image is a perfect match to its filter and the other 
has zero overlap with its filter (\emph{the motion in question is not present}). 
The situation can only become worse for feature vectors that account for less than $50\%$ 
of the variability.

\subsection{Synchrony detection using multiplicative interactions}
If one is willing to abandon weighted sums as the only allowable type of module 
for constructing deep networks, then a simple way to detect synchrony 
is by allowing for multiplicative (``gating'') interactions between filter responses: 
The product 
\begin{equation}
p=\vec{w}_2^\mathrm{T}\vec{x}_2 \cdot \vec{w}_1^\mathrm{T}\vec{x}_1
\end{equation}
will be large 
only if 
$\vec{w}_2^\mathrm{T}\vec{x}_2$ and $\vec{w}_1^\mathrm{T}\vec{x}_1$ 
both take on large (or both very negative) values. 
Any sufficiently small response of either 
$\vec{w}_2^\mathrm{T}\vec{x}_2$ or $\vec{w}_1^\mathrm{T}\vec{x}_1$ will shut off
the response of $p$, regardless of the filter response on the other image. 
That way, even a low threshold on $p$ will not sacrifice our ability to differentiate 
between the presence of some feature in one of the images vs. the \emph{partial} presence 
of the transformed feature in both of the images (synchrony). 

A related, less formal, argument for product interactions 
is that synchrony detection amounts to an operation akin to a logical ``AND''.  
This is at odds with the observation that weighted sums ``accumulate'' information 
and resemble a logical ``OR'' rather than an ``AND'' \citep[e.g.][]{zetzsche2005}.

\subsection{A locally learned gating module}
\begin{figure}[ht]
\vskip 0.2in
\begin{center}    
    \includegraphics[scale=0.3]{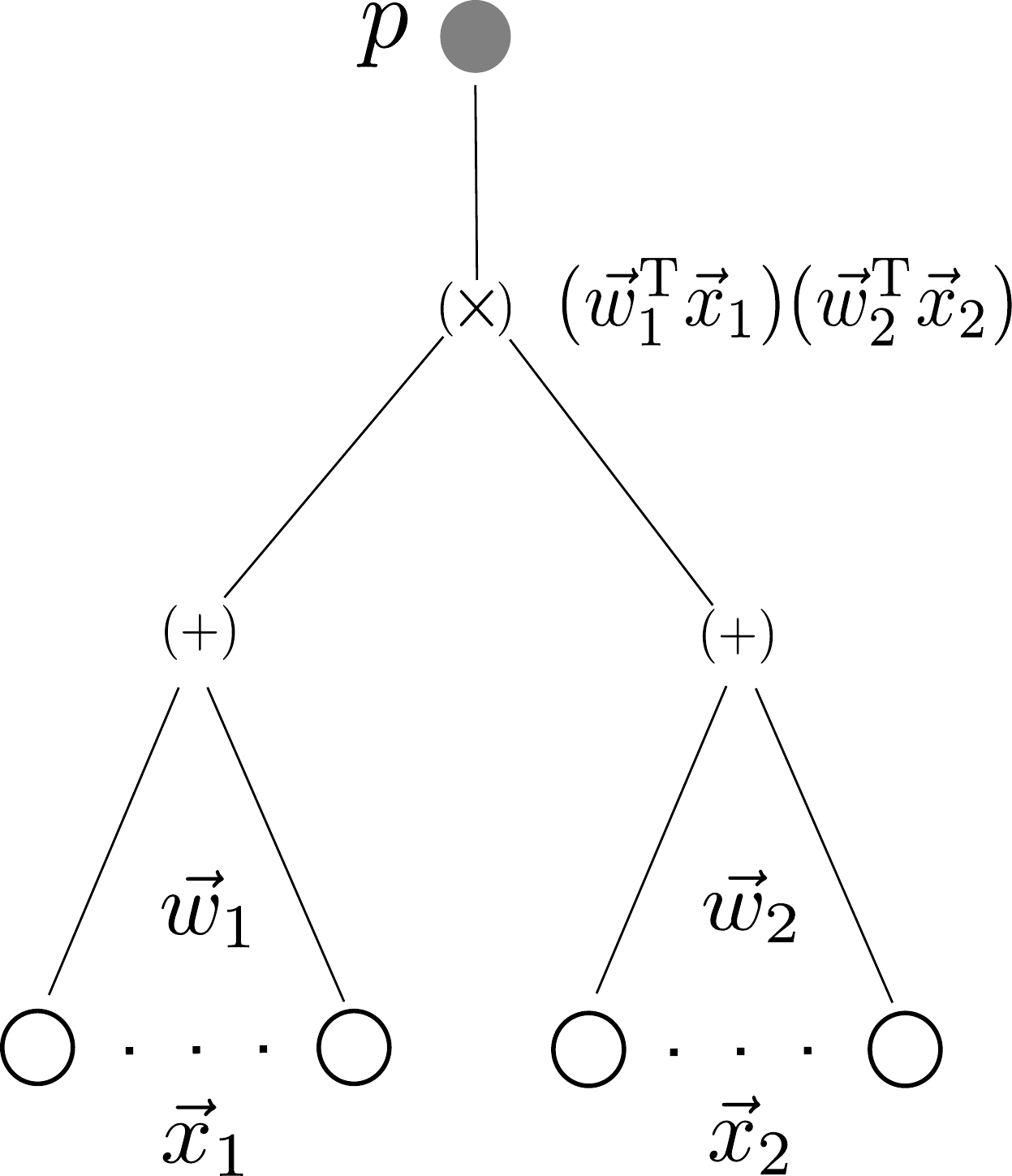}
    \caption{Gating within a ``dendritic tree.''}
    \label{figure:dendriticgating}
\end{center}
\vskip -0.2in
\end{figure}

It is important to note that multiplicative interactions will allow us to check 
for the synchrony condition using entirely \emph{local} operations: 
Figure~\ref{figure:dendriticgating} illustrates how we may define a ``neuron''  
that can detect synchrony by allowing for gating interactions within its ``dendritic tree''. 
A model consisting of multiple of these synchrony detector units will be a single-layer 
model, as there is no cross-talk required between the units. 
As we shall show, this fact allows us to use very fast local update rules for learning 
synchrony from data. 

This is in stark contrast to the learning of energy models and bi-linear models 
\citep[e.g.][]{GrimesRao,HyvarinenISA,Memisevic07,bethge2007,Taylor:2010},
which rely on non-local 
computations, such as back-prop, for learning (see also, Section~\ref{subsection:pooling}). 
Although multiplicative interactions have been a common ingredient in most of these models 
their motivation has been that they allow for the computation of subspace energies 
or subspace angles rather than synchrony \citep[eg.][]{multiview}.  

The usefulness of intra-dendritic gating has been discussed at lengths in the 
neuroscience literature, for example, in the work by Mel and 
colleagues \citep[e.g.][]{archiemel2000,Mel94}. But besides multi-layer bilinear models 
discussed above, it has not received much attention in machine learning. 
Dendritic gating is reminiscent also of 
``Pi-Sigma'' neurons \cite{Shin91thepi-sigma}, which have been applied to some 
supervised prediction tasks in the past. 

\subsection{Pooling and energy models}
\label{subsection:pooling}
\begin{figure*}[ht]
\vskip 0.2in
\begin{center}    
 \tabcolsep=1pt
 \begin{tabular}{|c c c c c c|}
 \hline
  &  & & case-1 & case-2 & case-3 \\
  $\vec{w}_1$ & \raisebox{-.5\height}{\includegraphics[scale=0.13]{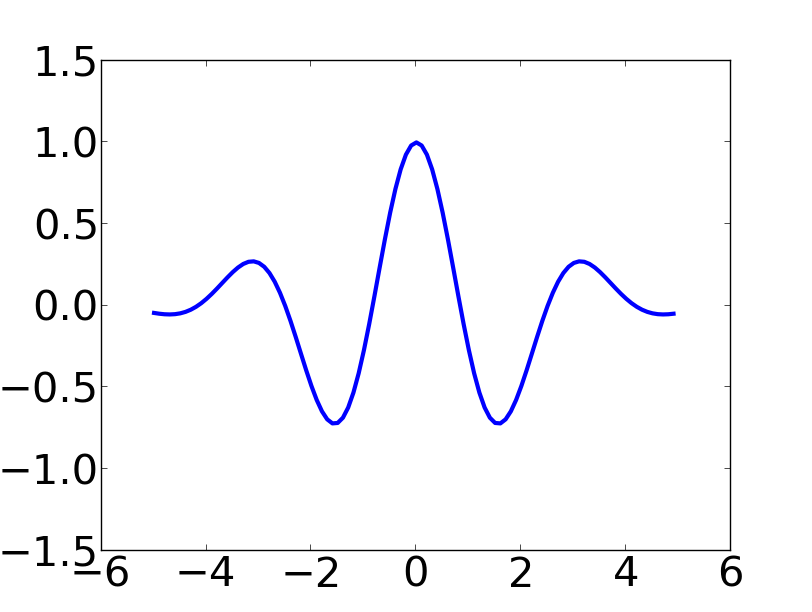}} & $\vec{x}_1$ & \raisebox{-.5\height}{\includegraphics[scale=0.13]{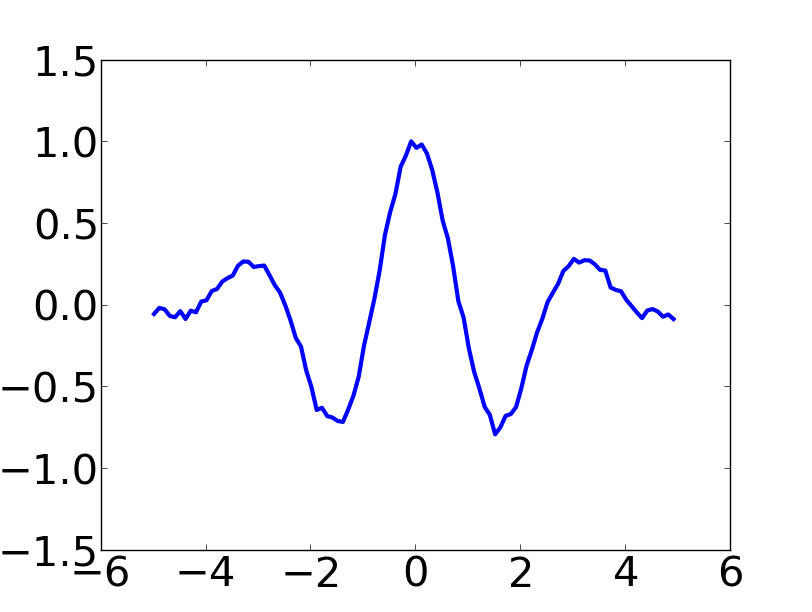}} & \raisebox{-.5\height}{\includegraphics[scale=0.13]{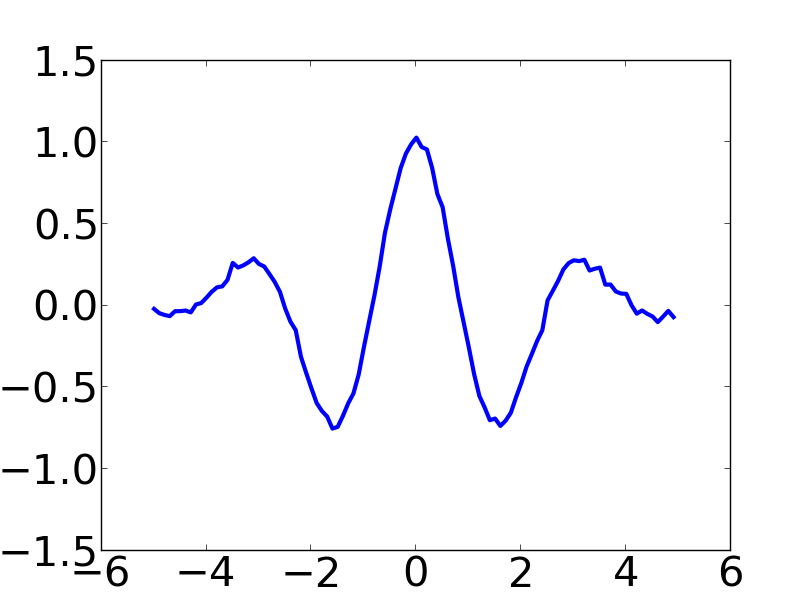}} & \raisebox{-.5\height}{\includegraphics[scale=0.13]{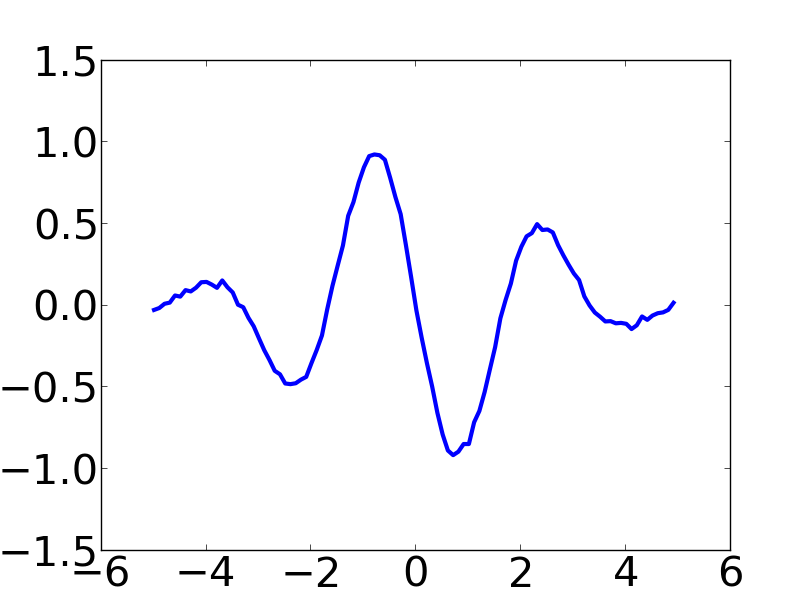}} \\
  $\vec{w}_2$ & \raisebox{-.5\height}{\includegraphics[scale=0.13]{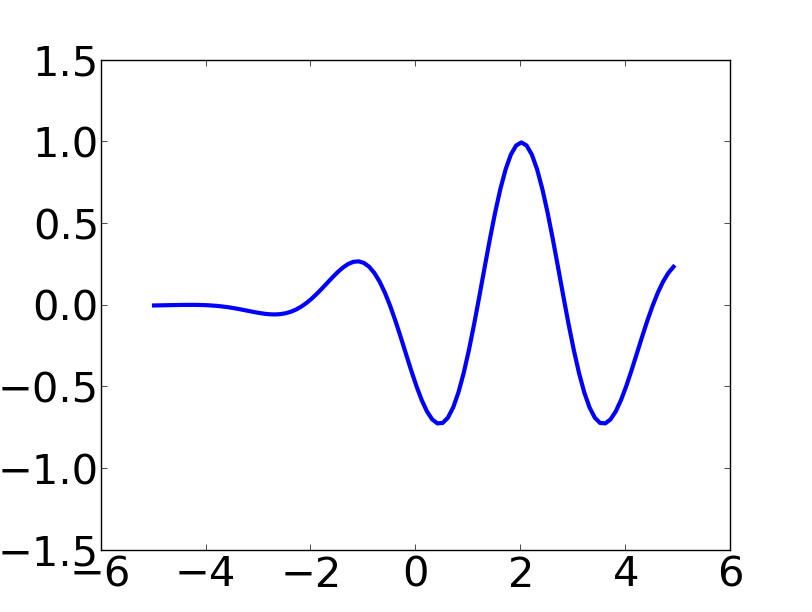}} & $\vec{x}_2$ & \raisebox{-.5\height}{\includegraphics[scale=0.13]{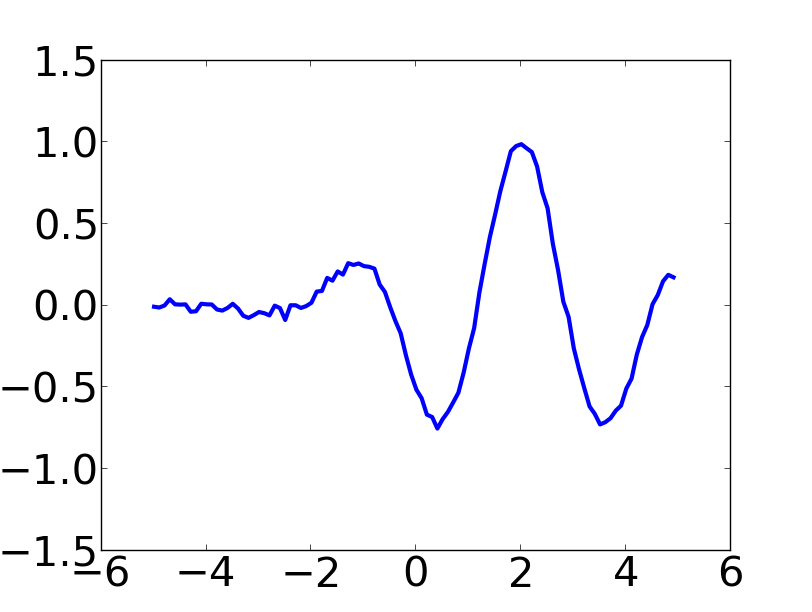}} & \raisebox{-.5\height}{\includegraphics[scale=0.13]{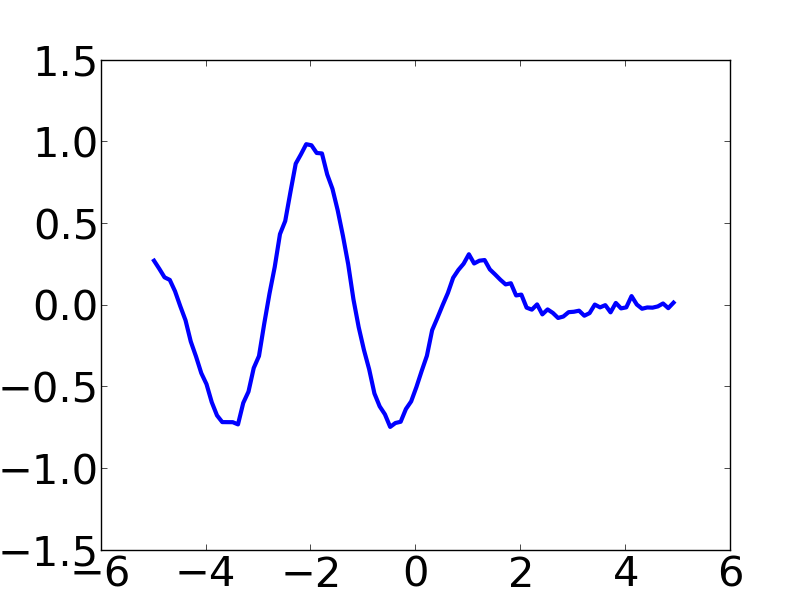}} & \raisebox{-.5\height}{\includegraphics[scale=0.13]{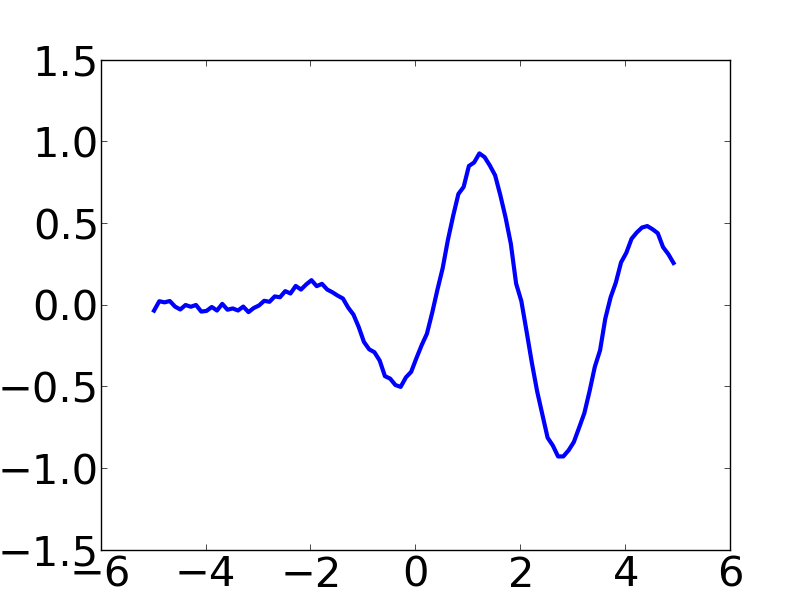}} \\
  & & & & & \\
  & & & $p=1231.02$ & $p=37.52$ & $p=-0.002$ \\
  \hline
 \end{tabular}
 \caption{Demonstration of product responses $p$ with two filters $\vec{w}_1, \vec{w}_2$ 
encoding a translation $P$. 
case-1: $\vec{x}_2 = P\vec{x}_1$; 
case-2: $\vec{x}_2 \neq P\vec{x}_1$; 
case-3: $\vec{x}_2 = P\vec{x}_1$ but $\vec{x}_1$ and $\vec{w}_1$ are out of phase by $\pi/2$.}
\label{productdemo}
\end{center}
\vskip -0.2in
\end{figure*}

Figure~\ref{productdemo} shows an illustration of a product response using a 1-D example.  
The figure shows how the product of transformed filters and inputs yields a 
large response whenever (i) the input is well-represented by the filter and 
(ii) the input evolves over time in a similar way as the filter (second column in the figure). 
The figure also illustrates how failing to satisfy either (i) or (ii) 
will yield a small product response (two rightmost columns). 
The need to satisfy condition (i) makes the product response dependent 
on the input. This dependency can be alleviated by 
\emph{pooling} over multiple products, involving multiple different filters, such that 
the top-level pooling unit fires, if any subset of the synchrony detectors fires. 
The classic energy model, for example, pools over filter pairs in quadrature to 
eliminate the dependence on \emph{phase} \cite{Adelson85,FleetBinocular}. 
In practice, however, it is not just phase but also frequency, position and orientation 
(or entirely different properties for non-Fourier features), 
which will determine whether an image is aligned with a filter or not. 
We investigate pooling with a separately trained pooling layer in Section~\ref{section:model}.

\section{Learning synchrony from data}
\label{section:model}
We now discuss how to learn filters which allow us to detect the synchrony condition. 
There are in principle many ways to achieve this in practice, 
and we introduce a temporal variant of the K-means algorithm to learn synchrony.
In Appendix~\ref{section:Synchronyautoencoder} we present another model 
based on the contractive autoencoder \cite{contractiveAE}, which we call synchrony autoencoder (SAE).

In the following, we let $\vec{x}, \vec{y} \in\mathbb{R}^{N}$ denote images, 
and we let $\mathbf{W}^x, \mathbf{W}^y \in\mathbb{R}^{Q\times N}$ denote matrices whose 
rows contain $Q$ feature vectors, which we will denote 
by $\vec{W}^x_q, \vec{W}^y_q \in\mathbb{R}^{N}$.  

\subsection{Synchrony K-means}
\label{subsection:synchronykmeans}
Online K-means clustering has recently been shown to yield efficient,  
and highly competitive image features for objective recognition \cite{coatessinglelayer}. 

We first note that, given a set of $Q$ cluster centers $\vec{W}^x_q$, performing 
online gradient-descent on the standard (not synchrony) K-means clustering objective
is equivalent to updating the cluster centers using the local competitive 
learning rule \cite{RumelhartZipser}
\begin{equation}
\Delta \vec{W}^x_s = \eta (\vec{x}-\vec{W}^x_s) 
\end{equation}
where $\eta$ is a step-size and $s$ is the ``winner-takes-all'' assignment
\begin{equation}
 s = \text{arg min}_q \|\vec{x}-\vec{W}^x_q\|^2
\end{equation}
When cluster-centers (``features'') are contrast-normalized,  
the assignment function is equivalent to
\begin{equation}
s = \text{arg max}_q [(\vec{W}^x_q)^\mathrm{T}\vec{x}]
\end{equation}

With the online K-means rule in mind, we now define a synchrony K-means (SK-means) model 
as follows.
We define the synchrony condition by first introducing multiplicative interactions 
in the assignment function:  
\begin{equation}
 s = \text{arg max}_q [((\vec{W}_q^x)^\mathrm{T}\vec{x})((\vec{W}_q^y)^\mathrm{T}\vec{y})]
\end{equation}
Note that computing the multiplication 
is equivalent to replacing the K-means winner-takes-all units by gating 
units (cf., Figure~\ref{figure:dendriticgating}).
This allows us to redefine the K-means objective function to be the reconstruction error between 
one input and the assigned prototype vector, which is gated (multiplied elementwise) with 
the projection of the other input:  
\begin{equation}
L_x = ( \vec{x} - \vec{W}^x_s ((\vec{W}_s^y)^\mathrm{T}\vec{y}))^2
\end{equation}
The gradient of the reconstruction error is 
\begin{equation}
 \frac{\partial L_x}{\partial \vec{W}^x_s} 
    = -4 (\vec{x}(\vec{W}^y_s)^\mathrm{T}\vec{y} - \vec{W}^x_s ((\vec{W}^y_s)^\mathrm{T}\vec{y})^2 )
   \label{gradient_sk}
\end{equation}
This allows us to define the \emph{synchrony K-means learning rule}: 
\begin{equation}
 \Delta \vec{W}^x_s = \eta (\vec{x}(\vec{W}^y_s)^\mathrm{T}\vec{y} - \vec{W}^x_s ((\vec{W}^y_s)^\mathrm{T}\vec{y})^2 )
\end{equation}
Similar to the online-kmeans rule \cite{RumelhartZipser}, 
we obtain a Hebbian term $\vec{x}(\vec{W}^y_s)^\mathrm{T}\vec{y}$, and an ``active forgetting'' term 
$(\vec{W}^x_s ((\vec{W}^y_s)^\mathrm{T}\vec{y})^2)$
which enforces competition among the hiddens. 
The Hebbian term, in contrast to standard K-means, is ``gated'', in that it 
involves both the ``pre-synaptic'' input $\vec{x}$, and the 
projected pre-synaptic input $(\vec{W}^y_s)^\mathrm{T}\vec{y}$ coming from the other input. 
Similarly the update rule for $\vec{W^y_s}$ is given by 
\begin{equation}
 \Delta \vec{W}^y_s = \eta (\vec{y}(\vec{W}^x_s)^\mathrm{T}\vec{x} - \vec{W}^y_s ((\vec{W}^x_s)^\mathrm{T}\vec{x})^2 )
\end{equation}

\begin{figure*}
\begin{center}
  \subfigure{\includegraphics[width=0.16\linewidth]{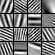}} \hspace{3mm}
  \subfigure{\includegraphics[width=0.16\linewidth]{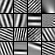}} \hspace{3mm}
  \subfigure{\includegraphics[width=0.16\linewidth]{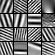}} \hspace{3mm}    
  \subfigure{\includegraphics[width=0.16\linewidth]{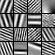}} \hspace{5mm} 
  \subfigure{\includegraphics[width=0.188\linewidth]{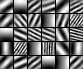}} \\

  \subfigure{\includegraphics[width=0.16\linewidth]{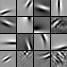}} \hspace{3mm}
  \subfigure{\includegraphics[width=0.16\linewidth]{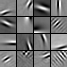}} \hspace{3mm}
  \subfigure{\includegraphics[width=0.16\linewidth]{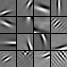}} \hspace{3mm}
  \subfigure{\includegraphics[width=0.16\linewidth]{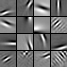}} \hspace{5mm} 
  \subfigure{\includegraphics[width=0.188\linewidth]{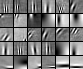}}
  \caption{{\bf Row 1:} Filters learned on synthetic translations of natural image patches. 
{\bf Row 2:} Filters learned on natural videos. 
{\bf Columns 1-4:} Frames 1-4 of the learned filters. 
  {\bf Column 5:} Filter groupings learned by a separate layer of K-means 
(only first frame filters shown). Each row in column 5 shows the six filters 
contributing the most to that cluster center.}
  \label{filters_grouping}
\end{center}
\end{figure*}

\subsection{Synchrony detection using even-symmetric non-linearities}
\label{subsection:evensymmetric}
As defined in Section \ref{subsection:synchrony_detection}, $\vec{x}_i, \vec{w}_i \; (i=1,\dots,T)$ denote the input 
frames and corresponding filters. An even-symmetric nonlinearity with global minimum at zero, such as the 
square function, applied to $\sum_i \vec{w}_i^\mathrm{T} \vec{x}_i$, will be a 
detector of the synchrony condition, too. 
The reason is the binomial identity, which states that the square of the sum of terms 
contains the pair-wise products between all individual terms plus the squares of the individual terms. 
The latter do not change the preferred stimulus of the unit \cite{FleetBinocular,multiview}.
The value of $(\sum_i \vec{w}_i^\mathrm{T} \vec{x}_i)^2$ is high only when the individual terms
are equal to each other and of high value, i.e, $\vec{w}_i^\mathrm{T} \vec{x}_i = \vec{w}_j^\mathrm{T} \vec{x}_j$
which is the synchrony condition in case of sequences (Equation \ref{synchrony_seq}).
Squaring non-linearities applied to the sum of phase-shifted Gabor filter responses 
have been the cornerstone of the energy model \cite{Adelson85,Watson:85,Hyvarinen}. 

Even-symmetric non-linearities implicitly compute pair-wise products and 
they may be implemented using multiplicative interactions, too: 
Consider the unit in Figure~\ref{figure:dendriticgating}, using ``tied'' inputs 
$\vec{x}_1=\vec{x}_2$, and assume that they contain a video sequence rather than a single image. 
If we also use tied weights $\vec{w}_1=\vec{w}_2$, then the output, $p$, of this unit 
will be equal to the square of $\vec{w}_1^\mathrm{T}\vec{x}_1$. 
In practice, the model can learn to tie weights, if required. 

To enable the model from Section \ref{subsection:synchronykmeans}
to encode motion across multiple frames, we may thus proceed as follows: 
Let $\vec{X}\in\mathbb{R}^{N}$ be 
the concatenation of $T$ frames $\vec{x}_{t}\in\mathbb{R}^{M}, t=1,\dots,T$, and 
let $\mathbf{W}\in\mathbb{R}^{Q\times N}$ denote the matrix containing 
the $Q$ feature vectors $\vec{W}_q\in\mathbb{R}^{N}$ stacked row-wise.
Each feature is composed of frame features $\vec{w}_{qt}\in\mathbb{R}^{M}$ 
where each $\vec{w}_{qt}$ spans one frame $\vec{x}_{t}$ from the input video. 

The SK-means can be adapted to sequences by replacing frames
$\vec{x}, \vec{y}$ with a sequence $\vec{X}$ and tying the weights $\mathbf{W^x}, \mathbf{W^y}$ to $\mathbf{W}$. 
The update rule for the SK-means model now becomes
\begin{equation}
 \Delta \vec{W}_s = \eta (\vec{X}(\vec{W}_{s}^\mathrm{T}\vec{X}) - \vec{W}_{s} (\vec{W}_{s}^\mathrm{T}\vec{X})^2 )
\end{equation}
where the assignment function $s$ is 
\begin{equation}
\label{equation:wtasquare}
 s(\vec{X}) = \text{arg max}_q [(\vec{W}_q^\mathrm{T}\vec{X})^2]
\end{equation}
Note that computing the square of $\sum_t \vec{w}_{qt}^\mathrm{T}\vec{x}_t$ above 
also accounts for synchrony as explained earlier.

For inference in case of the SK-means model, we use a sigmoid activation function on the squared features in our experiments 
instead of winner-takes-all (cf., Eq.~\ref{equation:wtasquare}). 
As in the case of object classification \cite{coatessinglelayer}, 
relaxing the harsh sparsity induced by K-means tends to yield codes better 
suited for recognition tasks. 

Example filters learned with the contractive SAE on sequences
are shown in Figure~\ref{filters_grouping}.  
In the first row of the figure, columns $1$ to $4$ show filters learned on $50,000$ synthetic movies  
generated by translating image patches from the natural image dataset in \cite{berkeley}.
Columns $1$ to $4$ of the second row show filters learned on blocks sampled from
videos of a broadcast TV database in \cite{hat}.
We obtained similar filters using the SK-means model. 

\subsection{Learning a separate pooling layer}
To study the dependencies of features, we performed K-means clustering, 
using $500$ centroids, on the hiddens extracted from the training sequences. 
Column 5 of Figure~\ref{filters_grouping} shows, 
for the most active clusters across the training data, 
the six features which contribute the most to each of the cluster centers.   
It shows that the ``pooling units'' (cluster centers) group together features 
with similar orientation and position, and with arbitrary frequency and phase.  
This is to be expected, as translation in any direction will affect \emph{all} frequencies 
and phase angles, and only ``nearby'' orientations and positions. 
Note in particular, that pooling across phase angles alone, as done by the classic 
energy model, would not be sufficient, and it is, in fact, not the solution 
found by pooling.

\section{Application to activity recognition}
\label{section:experiments}
Activity recognition is a common task for evaluating models of motion understanding. 
To allow for a fair comparison, we use the same pipeline as described 
in \cite{ISA,Wang09evaluationof}, using the features learned by our models.
We train 
our models on pca-whitened input patches of size $10 \times 16 \times 16$. 
The number of training samples is $200,000$. The number of product units are fixed at $300$. 
For inference sub blocks of the same size as the patch size are cropped from ``super blocks''  
of size $14\times20\times20$ \cite{ISA}. 
The sub blocks are cropped with a stride of $4$ on each axis giving $8$ sub blocks 
per super block. The feature responses of sub blocks are concatenated and dimensionally reduced using PCA to form the local feature.
Using a separate layer of K-means, a vocabulary of $3000$ spatio-temporal words is learned with $500,000$ 
samples for training. 
In all our experiments the super blocks are cropped densely from the video with a $50\%$ overlap.
Finally, a $\chi^{2}$-kernel SVM on the histogram of spatio-temporal words 
is used for classification. 

\subsection{Datasets}
We evaluated our models on several popular activity recognition benchmark datasets: 

KTH \cite{1334462}: Six actions performed by $25$ subjects. Samples divided into train and test data according to the authors original split.
The multi-class SVM is directly used for classification. 

UCF sports\cite{Rodriguez08actionmach:}: Ten action classes. The total number of videos in the dataset is 150. To increase the data we add horizontally flipped version of each video to the dataset. Like in \cite{Rodriguez08actionmach:} we train a multi-class SVM for classification, and we use leave-one-out for evaluation. That is, each original video is tested with all other videos as training set except the flipped version of the one being tested and itself.

Hollywood2 \cite{marszalek09}: Twelve activity classes. 
It consists of 884 test samples and 823 train samples with some of the video samples belonging to multiple classes. Hence, a binary SVM is used to compute the average precision (AP) of each class and the mean AP over all classes is reported \cite{marszalek09}.

YUPENN dynamic scenes \cite{YUPENN}: 
Fourteen scene categories with 30 videos for each category. We only use the gray-scale version of the videos in our experiments. Leave-one-out cross-validation is used for performance evaluation \cite{YUPENN}.

\begin{table}
\caption{Average accuracy on KTH.}
\label{kth}
\vskip 0.15in
\begin{center}
\begin{small}
\begin{sc}
\begin{tabular}{l c}
 \hline
 \abovespace\belowspace
 Algorithm 		& Performance(\%) \\
 \hline
 \abovespace
 {\bf SAE}	 	& 93.5 \\
 {\bf SK-means}		& 93.6 \\
 GRBM\cite{Taylor:2010}	& 90.0 \\
 \belowspace
 ISA model\cite{ISA}	& 93.9 \\ 
 \hline
\end{tabular}
\end{sc}
\end{small}
\end{center}
\vskip -0.1in
\end{table}

\begin{table}
\caption{Average accuracy on UCF sports.}
\label{ucf}
\vskip 0.15in
\begin{center}
\begin{small}
\begin{sc}
\begin{tabular}{l c}
 \hline
 \abovespace\belowspace
 Algorithm 		& Performance(\%) \\
 \hline
 \abovespace
 {\bf SAE}		& 86.0 \\
 {\bf SK-means}		& 84.7 \\
 \belowspace
 ISA model\cite{ISA}	& 86.5 \\
 \hline
\end{tabular}
\end{sc}
\end{small}
\end{center}
\vskip -0.1in
\end{table}

\begin{table}
\caption{Mean AP on Hollywood2.}
\label{holly}
\vskip 0.15in
\begin{center}
\begin{small}
\begin{sc}
\begin{tabular}{l c}
 \hline
 \abovespace\belowspace
 Algorithm 		& Performance(\%) \\
 \hline
 \abovespace
 {\bf SAE}	 	& 51.8 \\
 {\bf SK-means}		& 50.5 \\
 GRBM\cite{Taylor:2010}	& 46.6 \\ 
 ISA model\cite{ISA}	& 53.3 \\ 
 \belowspace
 covAE \cite{gated}	& 43.3 \\ 
 \hline
\end{tabular}
\end{sc}
\end{small}
\end{center}
\vskip -0.1in
\end{table}

\begin{table}
\caption{Average accuracy on YUPENN.}
\label{yupenn}
\vskip 0.15in
\begin{center}
\begin{small}
\begin{sc}
\begin{tabular}{l c}
 \hline
 \abovespace\belowspace
 Algorithm 		& Performance(\%) \\
 \hline
 \abovespace
 {\bf SAE} (k-NN)		& 80.7 \\
 {\bf SAE} ($\chi^2$svm)	& 96.0 \\
 {\bf SK-means} ($\chi^2$svm)	& 95.2 \\
 \belowspace
 SOE \cite{YUPENN}	 	& 79.0 \\ 
 \hline 
\end{tabular}
\end{sc}
\end{small}
\end{center}
\vskip -0.1in
\end{table}

\subsection{Results}
The results are shown in Tables~\ref{kth}, \ref{ucf}, \ref{holly} and \ref{yupenn}. 
They show that the SAE and SK-means are competitive with the state-of-the-art, although 
learning is simpler than for most existing methods. 
To evaluate the importance of element-wise products of hidden units, we also evaluated K-means 
as well as a standard autoencoder with contraction as regularization on the Hollywood2 dataset.
The models achieved an average precision of $42.1$ and $42.7$ respectively, which is much lower 
than the performance from SAE and SK-means.
We also tested the covariance auto-encoder \cite{gated}, which learns  
an additional mapping layer that pools over squared simple cell responses.
Table \ref{holly} shows that the performance of this model is also considerably lower than 
our single-layer models, showing that learning the pooling layer along with features did not help. 

\subsection{Unsupervised learning and dataset bias}
To show that our models learn features that can generalize across datasets (``self-taught learning'' \cite{ISA}), 
we trained SAE on random samples from one of the datasets and used it for feature extraction 
to report performance on the others. 
The performances using the same metrics as before are shown 
in table \ref{crossvalidation}. 
It can be seen that the performance gets reduced by only a fairly small fraction as compared 
to training on samples from the respective dataset. 
Only in the case where training on the KTH dataset, performance 
on Hollywood2 is considerably lower. This is probably due to the less diverse 
activities in KTH as compared to those in Hollywood2. 

\begin{table}[t]
\caption{Performance on column dataset using SAE trained on row dataset.}
\label{crossvalidation}
\vskip 0.15in
\begin{center}
\begin{small}
\begin{sc}
\begin{tabular}{l c c c}
\hline
\abovespace\belowspace
Dataset		& KTH   & UCF	     & Hollywood2 \\
\hline
\abovespace
 KTH 			& 93.5 & 85.3      & 44.7      \\ 
 UCF	 		& 92.9 & 86.0      & 48.9      \\ 
 \belowspace
 Hollywood2    		& 92.7 & 85.3      & 51.8      \\ 
 \hline
\end{tabular}
\end{sc}
\end{small}
\end{center}
\vskip -0.1in
\end{table}

\begin{table}[t]
\caption{Training time.}
\label{timings}
\vskip 0.15in
\begin{center}
\begin{small}
\begin{sc}
\begin{tabular}{l c}
\hline
\abovespace\belowspace
Algorithm & Time \\
\hline
\abovespace
{\bf SK-means} (GPU)	& $2$ minutes  \\
{\bf SK-means} (CPU) 	& $3$ minutes  \\
{\bf SAE} (GPU) 	& $1-2$ hours  \\
ISA  \cite{ISA}	 	& $1-2$ hours  \\
\belowspace
GRBM \cite{Taylor:2010} & $2-3$ days \\
\hline
\end{tabular}
\end{sc}
\end{small}
\end{center}
\vskip -0.1in
\end{table}

\subsection{Computational efficiency}
Training times for learning the motion features 
are shown in Table \ref{timings}. They show 
that SK-means (trained on CPU) is orders of magnitude faster than 
all other models.
For the GPU implementations, we used the theano library \cite{bergstra+al:2010-scipy}.  
We also calculated inference times using a similar metric as \cite{ISA} and 
computed the time required to extract descriptors for 30 videos from the Hollywood2 
dataset with resolution $360\times 288$ pixels (with ``sigmoid-of-square'' hiddens
they are identical for SK-means or SAE). 
Average inference times (in seconds/frame) were $0.058$ on CPU and $0.051$ on GPU, 
making the models feasible in practical, and possibly real-time, applications.  
All experiments were performed on a system with a $3.20$GHz CPU, $24$GB RAM and a GTX 680 GPU.

\section{Conclusions}
Our work shows that learning about motion from videos can be simplified and significantly 
sped up by disentangling learning about the spatio-temporal evolution of the signal 
from learning about invariances in the inputs.  
This allows us to achieve competitive performance in activity recognition tasks
at a fraction of the computational cost for learning motion features required by 
existing methods, such as the motion energy model \cite{ISA}.  
We also showed how learning about motion is possible using entirely 
local learning rules. 

Computing products by using ``dendritic gating'' within individual, 
but competing, units may be viewed as an efficient compromise between 
bi-linear models, that are expensive because they encode interactions between all pairs 
of pixels \cite{GrimesRao,Memisevic07,OlshausenBilinear}, 
and ``factored'' models \citep[e.g.][]{OlshausesCadieu,Taylor:2010,multiview}, 
which are multi-layer models that rely on more complicated training schemes such as 
back-prop and which do not work as well for recognition. 


\textbf{Acknowledgments:}
This work was supported in part by the German Federal Ministry of Education and Research (BMBF) in 
project 01GQ0841 (BFNT Frankfurt), by an NSERC Discovery grant and by a Google faculty research award.

\appendix 

\section{Synchrony autoencoder}
\label{section:Synchronyautoencoder}
Here we present an additional approach to encoding motion across two frames, 
based on the contractive autoencoder \cite{contractiveAE}. 
Like the synchrony K-means algorithm, it can be extended to sequences with more 
than two frames, using an analogous construction (cf., Section~\ref{subsection:evensymmetric}).
Given two images, we first compute the linear filter 
responses $\vec{f}^x = \mathbf{W}^x\vec{x}$ and $\vec{f}^y = \mathbf{W}^y\vec{y}$. 
Given the derivations in Section~\ref{section:synchrony}, 
an encoding of the motion, $\vec{h}=\vec{h}(\vec{x}, \vec{y})$, inherent in the image sequence 
may then be defined as 
\begin{equation}
\label{2framesae}
  \vec{h} = \sigma(\vec{f}^x\odot \vec{f}^y)
\end{equation}
where $\odot$ is element-wise multiplication and $\sigma$ is the sigmoid nonlinearity $(1+\exp(-x))^{-1}$.
This definition makes sense only, if features vectors are related by the transformation we 
wish to detect. 
We shall now discuss how we can define a reconstruction criterion that enforces this 
criterion. 

The standard way to train an autoencoder on images is to add a decoder and to minimize 
reconstruction error.
In our case, because of the presence of multiplicative interactions in the encoder, 
the encoding loses information about the sign of the input. However, note that we may 
interpret the multiplicative interactions as gating as discussed in the previous section.
This suggests defining the reconstruction error on one input, given the other.  
In the decoder we thus perform an element-wise multiplication of the hiddens and factors of one of the input 
to reconstruct the other. One may also view this as re-introducing the sign information 
at reconstruction time. Assuming an autoencoder with tied weights, the reconstructed inputs can then be defined as 
\begin{eqnarray}
\label{decoder}
\hat{x}= (\mathbf{W}^x)^\mathrm{T}(\vec{h} \odot \vec{f}^y) \\
\hat{y}= (\mathbf{W}^y)^\mathrm{T}(\vec{h} \odot \vec{f}^x)
\end{eqnarray}
We define the reconstruction error as the average squared difference between the two inputs and 
their respective reconstructions: 
\begin{equation}
\label{reconstruction}
 L((\vec{x},\vec{y}),(\hat{\vec{x}},\hat{\vec{y}})) = \|(\vec{x} - \hat{\vec{x}})\|^2 + \|(\vec{y} - \hat{\vec{y}})\|^2  
\end{equation}
Learning amounts to minimizing the reconstruction error wrt. the filters $(\mathbf{W}^x)$ and $(\mathbf{W}^y)$.  
In contrast to bi-linear models, which may be trained using similar 
criteria \citep[e.g.][]{gated,Taylor:2010}, the representation of 
motion in Eq.~\ref{2framesae} will be dependent on the image content, such as Fourier 
phase for translational motion. 
But this dependence can be removed using a separately trained pooling layer as we 
shall show. The absence of pooling during feature learning allows for much more 
efficient learning as we show in Section~\ref{section:experiments}.
Note that, in practice, one may add bias terms to the definition of hiddens 
and reconstructions.

\subsection{Contractive regularization}
It is well-known that regularization is important to extract useful features 
and to learn sparse representations. 
Here, we use contraction as regularization \cite{contractiveAE}. 
This amounts to adding the Frobenius norm of the Jacobian of the extracted features, i.e., 
the sum of squares of all partial derivatives of $\vec{h}$ with respect to $\vec{x},\vec{y}$
\small
\begin{equation}
 \|J_e(\vec{x},\vec{y})\|^2_E = \sum_{ij}\left(\frac{\partial h_j(\vec{x},\vec{y})}{\partial x_i}\right)^2 
 + \sum_{ij}\left(\frac{\partial h_j(\vec{x},\vec{y})}{\partial y_i}\right)^2
\end{equation}
\normalsize
which for the sigmoid-of-square non-linearity becomes 
\small
\begin{eqnarray}
\label{contraction}
 \|J_e(\vec{x},\vec{y})\|^2_E & = \sum_{j}(h_j(1-h_j))^2 (f^x_j)^2 \sum_{i}(W^y_{ij})^2 \nonumber \\ 
		  & + \sum_{j}(h_j(1-h_j))^2 (f^y_j)^2 \sum_{i}(W^x_{ij})^2
\end{eqnarray}
\normalsize
For training, we add the regularization term to the reconstruction cost, 
using a hyperparameter $\lambda$. 
Contractive regularization is not possible in (multi-layer) bi-linear models,  
due to the computational complexity of computing the contraction gradient for 
multiple layers \citep[e.g.][]{gated}. 
Being a single layer model, the synchrony autoencoder (SAE) 
makes the application of contractive regularization feasible.
The contraction parameter $\lambda$ are set by cross-validation.


The SAE can be adapted to sequences by replacing frames
$\vec{x}, \vec{y}$ with a sequence $\vec{X}$ and tying the weights $\mathbf{W^x}, \mathbf{W^y}$ to $\mathbf{W}$. 
The representation of motion from Equation \ref{2framesae} can now be redefined as,
\begin{eqnarray}
H_q  =  \sigma(F_q^2)
     =  \sigma\Big(((\vec{W}_q)^\mathrm{T}\vec{X})^2\Big) 
     =  \sigma\Big((\sum_t \vec{w}_{qt}^\mathrm{T}\vec{x}_t)^2\Big) 
\label{hiddens}
\end{eqnarray}
Note that computing the square of $\sum_t \vec{w}_{qt}^\mathrm{T}\vec{x}_t$ above 
also accounts for synchrony as explained earlier. The reconstruction error 
and regularization term for this model can be derived by just replacing appropriate terms in
Equations \ref{reconstruction} and \ref{contraction}, respectively.

\bibliography{ref.bib}

\begin{thebibliography}{32}
\providecommand{\natexlab}[1]{#1}
\providecommand{\url}[1]{\texttt{#1}}
\expandafter\ifx\csname urlstyle\endcsname\relax
  \providecommand{\doi}[1]{doi: #1}\else
  \providecommand{\doi}{doi: \begingroup \urlstyle{rm}\Url}\fi

\bibitem[Adelson \& Bergen(1985)Adelson and Bergen]{Adelson85}
Adelson, Edward~H. and Bergen, James~R.
\newblock Spatiotemporal energy models for the perception of motion.
\newblock \emph{J. OPT. SOC. AM. A}, 2\penalty0 (2):\penalty0 284--299, 1985.

\bibitem[Archie \& Mel(2000)Archie and Mel]{archiemel2000}
Archie, Kevin~A. and Mel, Bartlett~W.
\newblock {A model for intradendritic computation of binocular disparity}.
\newblock \emph{Nature Neuroscience}, 3\penalty0 (1):\penalty0 54--63, January
  2000.

\bibitem[Arndt et~al.(1995)Arndt, Mallot, and B\"{u}lthoff]{arndt1995human}
Arndt, P.A., Mallot, H.A., and B\"{u}lthoff, H.H.
\newblock Human stereovision without localized image features.
\newblock \emph{Biological cybernetics}, 72\penalty0 (4):\penalty0 279--293,
  1995.

\bibitem[Bergstra et~al.(2010)Bergstra, Breuleux, Bastien, Lamblin, Pascanu,
  Desjardins, Turian, Warde-Farley, and Bengio]{bergstra+al:2010-scipy}
Bergstra, James, Breuleux, Olivier, Bastien, Fr{\'{e}}d{\'{e}}ric, Lamblin,
  Pascal, Pascanu, Razvan, Desjardins, Guillaume, Turian, Joseph, Warde-Farley,
  David, and Bengio, Yoshua.
\newblock Theano: a {CPU} and {GPU} math expression compiler.
\newblock In \emph{SciPy}, 2010.

\bibitem[Bethge et~al.(2007)Bethge, Gerwinn, and Macke]{bethge2007}
Bethge, M, Gerwinn, S, and Macke, JH.
\newblock Unsupervised learning of a steerable basis for invariant image
  representations.
\newblock In \emph{Human Vision and Electronic Imaging XII}. SPIE, 2007.

\bibitem[Cadieu \& Olshausen(2011)Cadieu and Olshausen]{OlshausesCadieu}
Cadieu, Charles~F. and Olshausen, Bruno~A.
\newblock {Learning Intermediate-Level Representations of Form and Motion from
  Natural Movies}.
\newblock \emph{Neural Computation}, 24\penalty0 (4):\penalty0 827--866,
  December 2011.

\bibitem[Coates et~al.(2011)Coates, Lee, and Ng]{coatessinglelayer}
Coates, Adam, Lee, Honglak, and Ng, A.~Y.
\newblock An analysis of single-layer networks in unsupervised feature
  learning.
\newblock In \emph{Artificial Intelligence and Statistics}, 2011.

\bibitem[Derpanis(2012)]{YUPENN}
Derpanis, Konstantinos~G.
\newblock Dynamic scene understanding: The role of orientation features in
  space and time in scene classification.
\newblock In \emph{CVPR}, 2012.

\bibitem[Fleet et~al.(1996)Fleet, Wagner, and Heeger]{FleetBinocular}
Fleet, D., Wagner, H., and Heeger, D.
\newblock {Neural encoding of binocular disparity: Energy models, position
  shifts and phase shifts}.
\newblock \emph{Vision Research}, 36\penalty0 (12):\penalty0 1839--1857, June
  1996.

\bibitem[Grimes \& Rao(2005)Grimes and Rao]{GrimesRao}
Grimes, David and Rao, Rajesh.
\newblock Bilinear sparse coding for invariant vision.
\newblock \emph{Neural Computation}, 17\penalty0 (1):\penalty0 47--73, 2005.

\bibitem[Hateren \& Schaaf(1998)Hateren and Schaaf]{hat}
Hateren, J. H.~van and Schaaf, A. van~der.
\newblock Independent component filters of natural images compared with simple
  cells in primary visual cortex.
\newblock \emph{Proceedings: Biological Sciences}, 265\penalty0
  (1394):\penalty0 359--366, Mar 1998.

\bibitem[Hyv\"{a}rinen \& Hoyer(2000)Hyv\"{a}rinen and Hoyer]{HyvarinenISA}
Hyv\"{a}rinen, Aapo and Hoyer, Patrik.
\newblock Emergence of phase- and shift-invariant features by decomposition of
  natural images into independent feature subspaces.
\newblock \emph{Neural Comput.}, 12:\penalty0 1705--1720, July 2000.

\bibitem[Hyvarinen et~al.(2009)Hyvarinen, Hurri, and Hoyer]{Hyvarinen}
Hyvarinen, Aapo, Hurri, Jarmo, and Hoyer, Patrick~O.
\newblock \emph{Natural Image Statistics: A Probabilistic Approach to Early
  Computational Vision}.
\newblock Springer Publishing Company, Incorporated, 2009.

\bibitem[Ji et~al.(2013)Ji, Xu, Yang, and Yu]{3DCNN}
Ji, Shuiwang, Xu, Wei, Yang, Ming, and Yu, Kai.
\newblock 3{D} convolutional neural networks for human action recognition.
\newblock \emph{IEEE Transactions on Pattern Analysis and Machine
  Intelligence}, 35\penalty0 (1):\penalty0 221--231, 2013.

\bibitem[Le et~al.(2011)Le, Zou, Yeung, and Ng]{ISA}
Le, Q.V., Zou, W.Y., Yeung, S.Y., and Ng, A.Y.
\newblock Learning hierarchical invariant spatio-temporal features for action
  recognition with independent subspace analysis.
\newblock In \emph{CVPR}, 2011.

\bibitem[Marsza{\l}ek et~al.(2009)Marsza{\l}ek, Laptev, and
  Schmid]{marszalek09}
Marsza{\l}ek, Marcin, Laptev, Ivan, and Schmid, Cordelia.
\newblock Actions in context.
\newblock In \emph{IEEE Conference on Computer Vision \& Pattern Recognition},
  2009.

\bibitem[Martin et~al.(2001)Martin, Fowlkes, Tal, and Malik]{berkeley}
Martin, D., Fowlkes, C., Tal, D., and Malik, J.
\newblock A database of human segmented natural images and its application to
  evaluating segmentation algorithms and measuring ecological statistics.
\newblock In \emph{ICCV}, 2001.

\bibitem[Mel(1994)]{Mel94}
Mel, Bartlett~W.
\newblock Information processing in dendritic trees.
\newblock \emph{Neural Computation}, 6\penalty0 (6):\penalty0 1031--1085, 1994.

\bibitem[Memisevic(2011)]{gated}
Memisevic, Roland.
\newblock Gradient-based learning of higher-order image features.
\newblock In \emph{ICCV}, 2011.

\bibitem[Memisevic(2012)]{multiview}
Memisevic, Roland.
\newblock On multi-view feature learning.
\newblock In \emph{ICML}, 2012.

\bibitem[Memisevic \& Hinton(2007)Memisevic and Hinton]{Memisevic07}
Memisevic, Roland and Hinton, Geoffrey.
\newblock Unsupervised learning of image transformations.
\newblock In \emph{CVPR}, 2007.

\bibitem[Olshausen(2003)]{OlshausenTimevarying}
Olshausen, B.A.
\newblock Learning sparse, overcomplete representations of time-varying natural
  images.
\newblock In \emph{Image Processing, 2003. ICIP 2003. Proceedings. 2003
  International Conference on}, volume~1, pp.\  I--41--4 vol.1, Sept 2003.

\bibitem[Olshausen et~al.(2007)Olshausen, Cadieu, Culpepper, and
  Warland]{OlshausenBilinear}
Olshausen, Bruno, Cadieu, Charles, Culpepper, Jack, and Warland, David.
\newblock Bilinear models of natural images.
\newblock In \emph{SPIE Proceedings: Human Vision Electronic Imaging
  {X}{I}{I}}, San Jose, 2007.

\bibitem[Rifai et~al.(2011)Rifai, Vincent, Muller, Glorot, and
  Bengio]{contractiveAE}
Rifai, Salah, Vincent, Pascal, Muller, Xavier, Glorot, Xavier, and Bengio,
  Yoshua.
\newblock {Contractive Auto-Encoders: Explicit Invariance During Feature
  Extraction}.
\newblock In \emph{ICML}, 2011.

\bibitem[Rodriguez et~al.(2008)Rodriguez, Ahmed, and
  Shah]{Rodriguez08actionmach:}
Rodriguez, Mikel~D., Ahmed, Javed, and Shah, Mubarak.
\newblock Action mach: a spatio-temporal maximum average correlation height
  filter for action recognition.
\newblock In \emph{CVPR}, 2008.

\bibitem[Rumelhart \& Zipser(1986)Rumelhart and Zipser]{RumelhartZipser}
Rumelhart, D.~E. and Zipser, D.
\newblock Parallel distributed processing: explorations in the microstructure
  of cognition, vol. 1.
\newblock chapter Feature discovery by competitive learning, pp.\  151--193.
  MIT Press, 1986.

\bibitem[Schuldt et~al.(2004)Schuldt, Laptev, and Caputo]{1334462}
Schuldt, C., Laptev, I., and Caputo, B.
\newblock Recognizing human actions: a local svm approach.
\newblock In \emph{Pattern Recognition, 2004. ICPR 2004. Proceedings of the
  17th International Conference on}, 2004.

\bibitem[Shin \& Ghosh(1991)Shin and Ghosh]{Shin91thepi-sigma}
Shin, Yoan and Ghosh, Joydeep.
\newblock The pi-sigma network: An efficient higher-order neural network for
  pattern classification and function approximation.
\newblock In \emph{International Joint Conference on Neural Networks}, 1991.

\bibitem[Taylor et~al.(2010)Taylor, Fergus, LeCun, and Bregler]{Taylor:2010}
Taylor, Graham~W., Fergus, Rob, LeCun, Yann, and Bregler, Christoph.
\newblock Convolutional learning of spatio-temporal features.
\newblock In \emph{Proceedings of the 11th European conference on Computer
  vision: Part VI}, ECCV'10, 2010.

\bibitem[Wang et~al.(2009)Wang, Ullah, Kl\"aser, Laptev, and
  Schmid]{Wang09evaluationof}
Wang, Heng, Ullah, Muhammad~Muneeb, Kl\"aser, Alexander, Laptev, Ivan, and
  Schmid, Cordelia.
\newblock Evaluation of local spatio-temporal features for action recognition.
\newblock In \emph{University of Central Florida, U.S.A}, 2009.

\bibitem[Watson \& Albert J.~Ahumada(1985)Watson and Albert
  J.~Ahumada]{Watson:85}
Watson, Andrew~B. and Albert J.~Ahumada, Jr.
\newblock Model of human visual-motion sensing.
\newblock \emph{J. Opt. Soc. Am. A}, 2\penalty0 (2):\penalty0 322--341, Feb
  1985.

\bibitem[Zetzsche \& Nuding(2005)Zetzsche and Nuding]{zetzsche2005}
Zetzsche, Christoph and Nuding, Ulrich.
\newblock {Nonlinear and higher-order approaches to the encoding of natural
  scenes.}
\newblock \emph{Network (Bristol, England)}, 16\penalty0 (2-3):\penalty0
  191--221, 2005.

\end{thebibliography}
\bibliographystyle{icml2014}

\end{document}